\documentclass[letterpaper, 10 pt, conference]{ieeeconf}  

\IEEEoverridecommandlockouts                              

\usepackage{graphics} 
\usepackage{epsfig} 
\usepackage{mathptmx} 
\usepackage{times} 
\usepackage{amsmath} 
\usepackage{amssymb}  
\usepackage[linesnumbered, ruled]{algorithm2e}
\usepackage[switch]{lineno}
\usepackage{booktabs}
\usepackage{threeparttable}
\usepackage{romannum}
\usepackage{soul}

\usepackage{color, xcolor}
\usepackage{here}
\usepackage{graphicx}
\usepackage{subfigure}
\usepackage{cite}
\usepackage{bm}
\usepackage{float}
\usepackage{mathtools}
\usepackage{amsmath,amsfonts}
\usepackage{multirow}
\renewcommand{\thesubsection}{\thesection.\Alph{subsection}}
\title{\LARGE \bf
A Fast Online Omnidirectional Quadrupedal Jumping Framework Via Virtual-Model Control and Minimum Jerk Trajectory Generation 
}

\author{Linzhu Yue$^{1}$, Lingwei Zhang$^{1}$, Zhitao Song$^{1}$, Hongbo Zhang$^{1}$,\\ Jinhu Dong$^{1}$, Xuanqi Zeng$^{1}$, and Yun-Hui Liu$^{1}$ 
\thanks{$^{1}$ L. Z. Yue, L. W. Zhang, Z. T. Song, H. B. Zhang, J. H. Dong, X. Q. Zeng, and Y.-H. Liu is with the Department of Mechanical and Automation Engineering at the Chinese University of Hong Kong.
        {\tt\small lzyue@mae.cuhk.edu.hk}}%
\thanks{* Corresponding author: Y.-H. Liu  
        {\tt\small yhliu@cuhk.edu.hk}}
\thanks{This work is supported by the InnoHK of the Government of Hong Kong via the Hong Kong Centre for Logistics Robotics, the CUHK T Stone Robotics Institute, and the Shenzhen Portion of Shenzhen-Hong Kong Science and Technology Innovation Cooperation Zone 
under HZQB-KCZYB-20200089.}
}

\begin{document}

\maketitle
\thispagestyle{empty}
\pagestyle{empty}

\begin{abstract}
Exploring the limits of quadruped robot agility, particularly in the context of rapid and real-time planning and execution of omnidirectional jump trajectories, presents significant challenges due to the complex dynamics involved, especially when considering significant impulse contacts. This paper introduces a new framework to enable fast, omnidirectional jumping capabilities for quadruped robots. Utilizing minimum jerk technology, the proposed framework efficiently generates jump trajectories that exploit its analytical solutions, ensuring numerical stability and dynamic compatibility with minimal computational resources. The virtual model control is employed to formulate a Quadratic Programming (QP) optimization problem to accurately track the Center of Mass (CoM) trajectories during the jump phase. The whole-body control strategies facilitate precise and compliant landing motion.
Moreover, the different jumping phase is triggered by time-schedule. The framework's efficacy is demonstrated through its implementation on an enhanced version of the open-source \textit{Mini Cheetah} robot. Omnidirectional jumps—including forward, backward, and other directional—were successfully executed, showcasing the robot's capability to perform rapid and consecutive jumps with an average trajectory generation and tracking solution time of merely 50 microseconds.


\end{abstract}

\section{INTRODUCTION}

Robotics jumping is critical for quadruped robots to traverse between complicated unstructured terrains. Although quasi-static jumping algorithms empower quadrupedal robots to cross mild irregular terrain, when aimed at avoiding unknown obstacles, it is necessary for robots to perform omnidirectional jumping promptly, which requires fast, skillful planning and dynamical executing motions. We strive to propose a framework with minimal computational resources that enables quadrupedal robots to make rapid omnidirectional jump responses to avoid sudden dangers while exploring unknown environments. This could be significantly helpful in inspection scenarios where quadrupedal robots are equipped with large, heavy instruments and the onboard computer has other heavy-load tasks. The real-time omnidirectional jumping framework could improve the robot's ability to traverse rough terrains and ensure the safety of robots.
\begin{figure}[!h]
\centering \label{Various_jump}
\vspace{0.25cm}
\hspace{-0.25cm}
 \subfigure[]{
    \label{jump_1} 
    \includegraphics[width=1.58in]{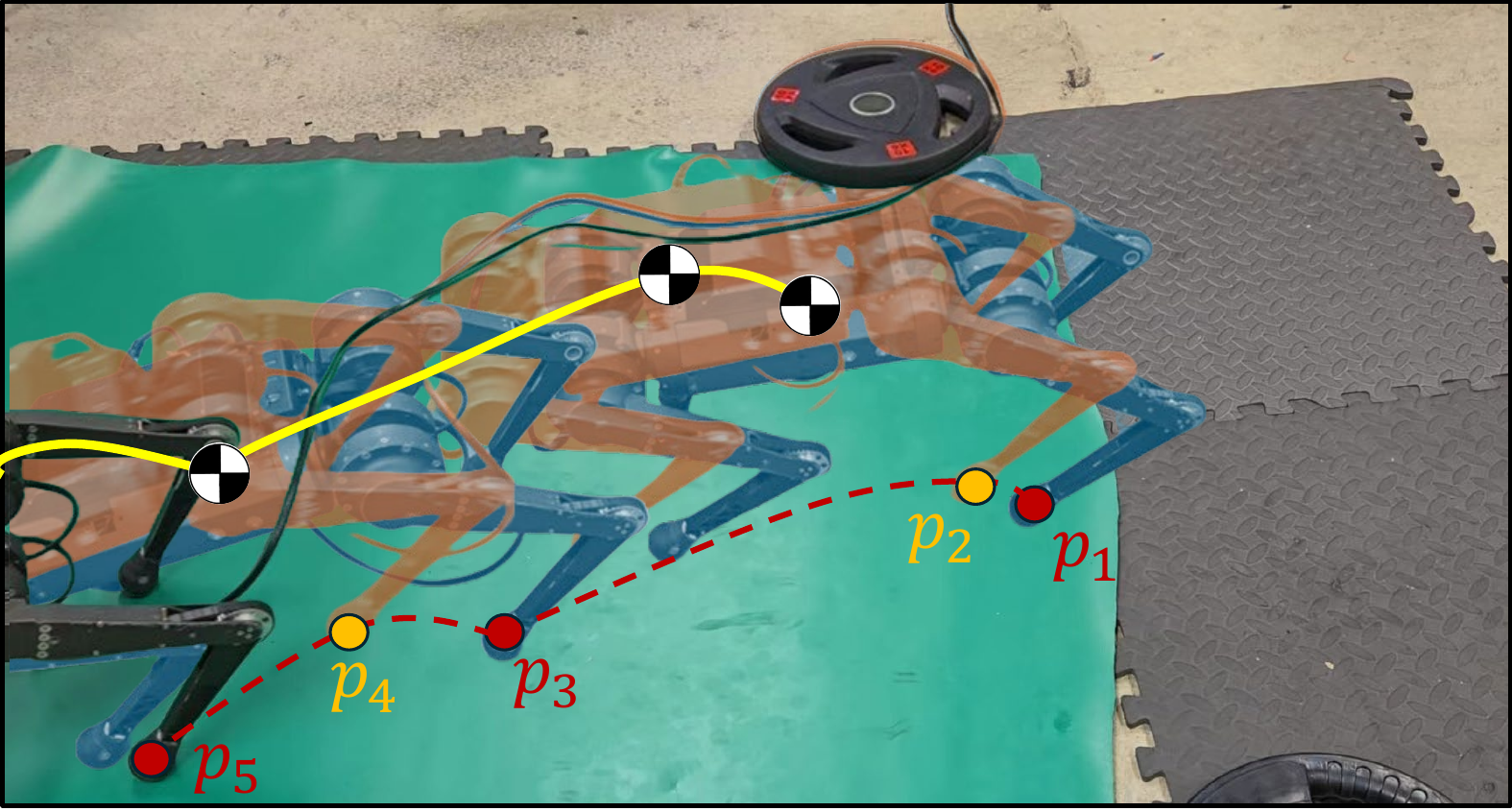}}
\subfigure[]{
    \label{jump_2} 
    \includegraphics[width=1.6in]{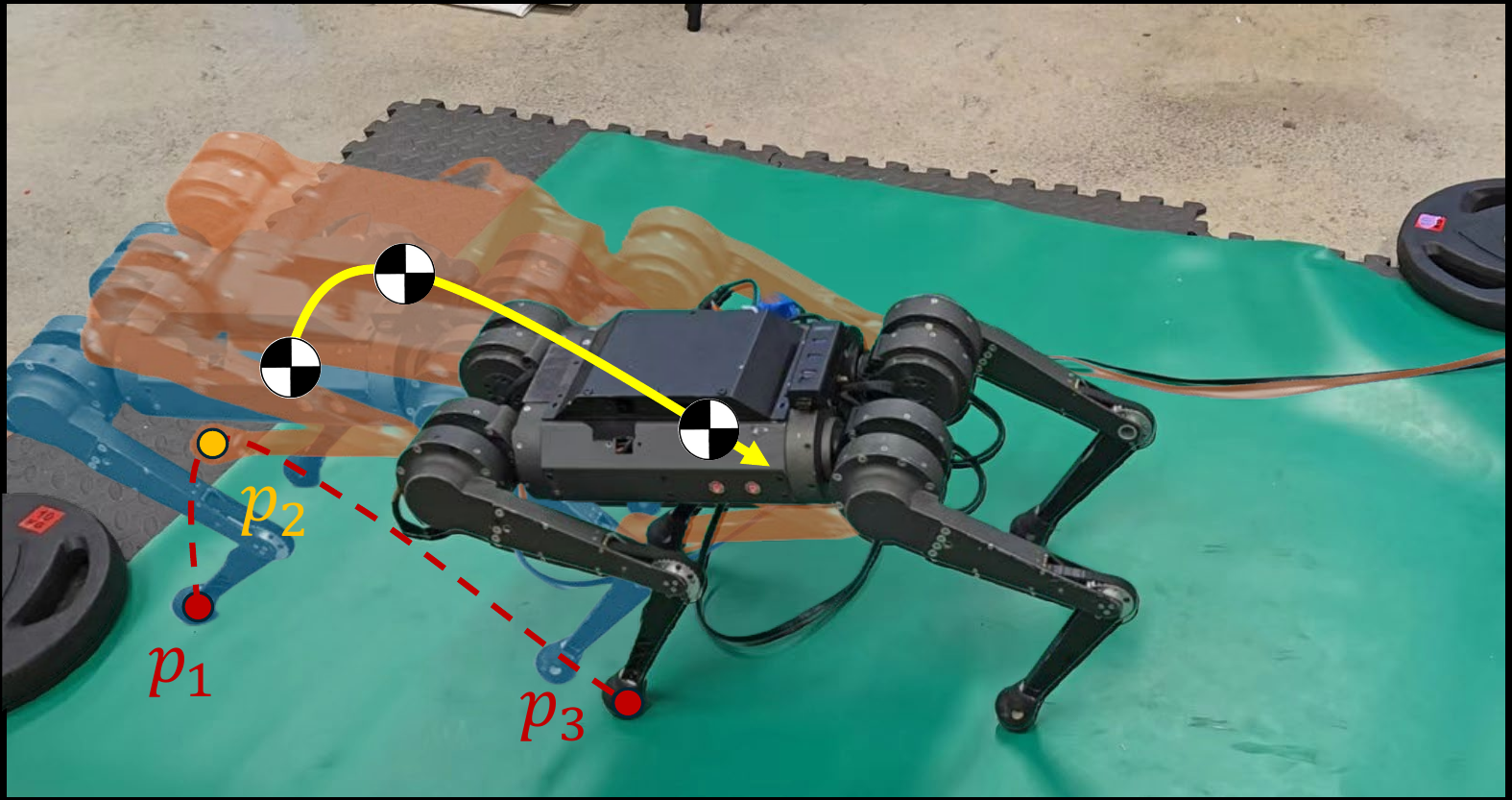}}
\subfigure[]{
    \label{jump_3} 
    \includegraphics[width=1.6in]{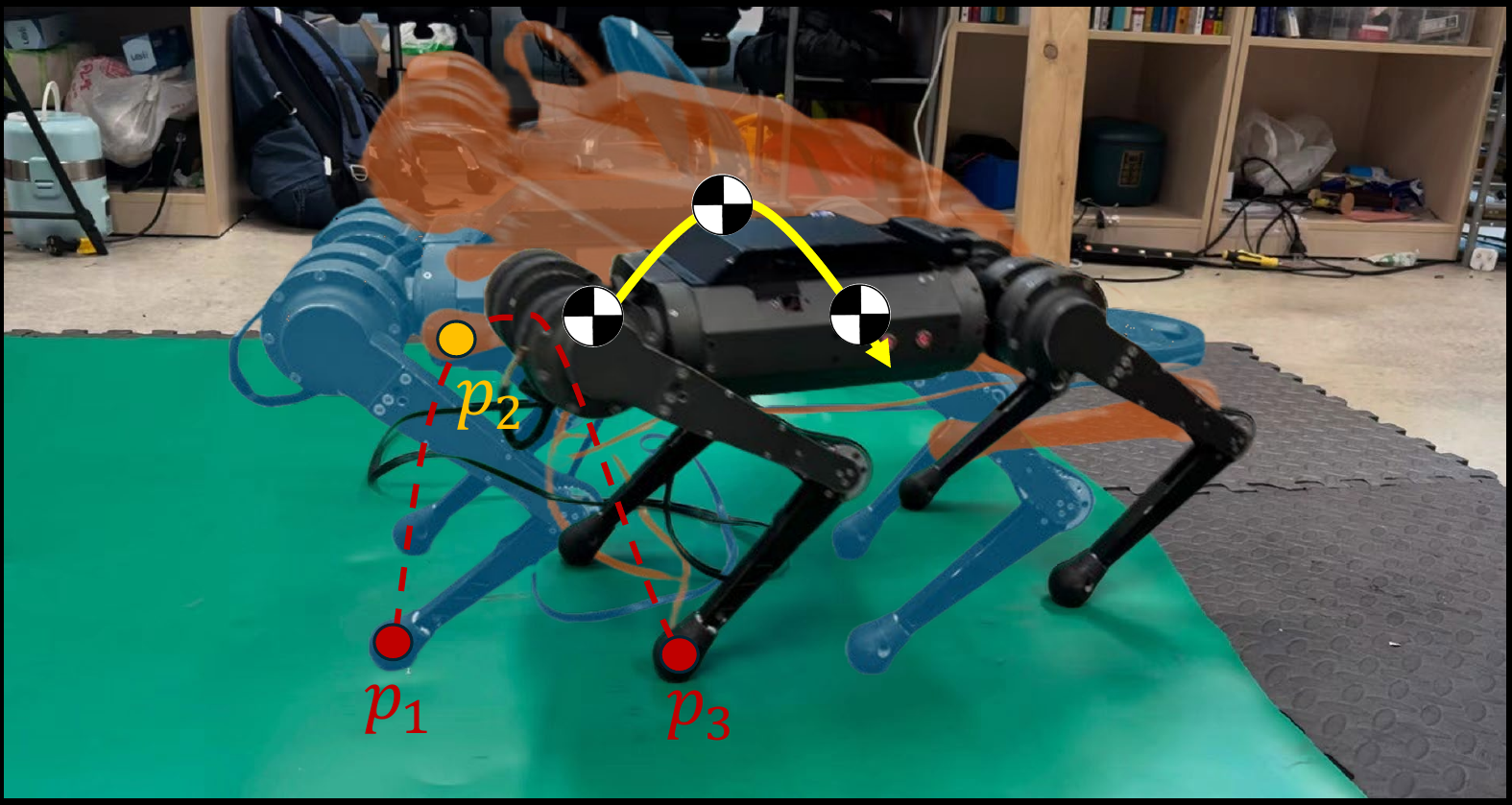}}
\subfigure[]{
    \label{jump_4} 
    \includegraphics[width=1.6in]{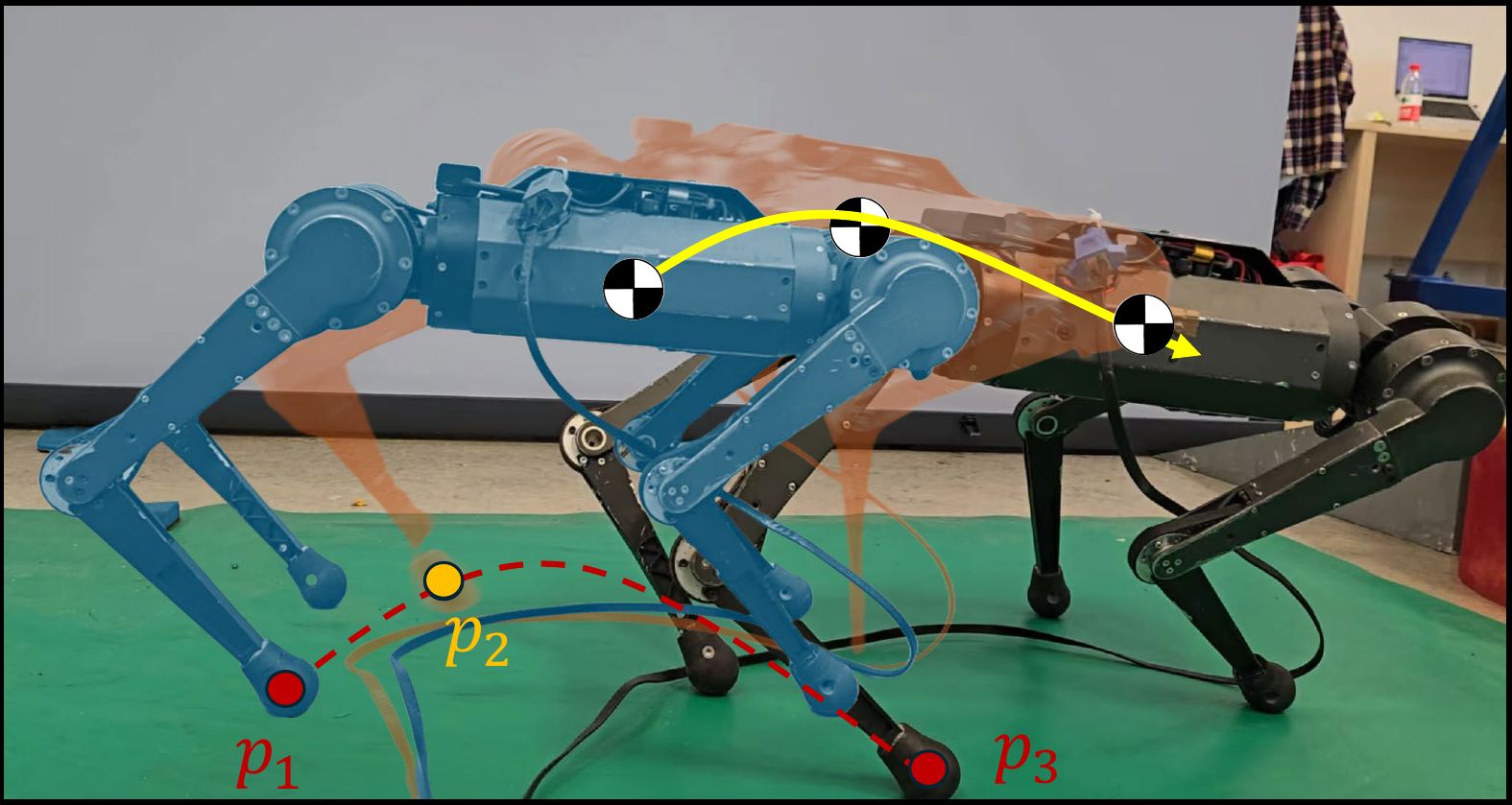}}
 \vspace{-0.2cm}
\caption{Various jumping motion experiments to validate the proposed omnidirectional jumping framework. (a) Front-left jumping consecutively. (b) Single rear-right jumping. (c) Single rear jumping. (d) Single front jumping. The yellow line shows the trajectory of CoM, and the red dashed line represents the trajectory of the selected foot. The red dots indicate the foot's contact with the ground, and the orange dots indicate the foot in the flight phase.}
\end{figure}

Various experimental conductions demonstrate the great potential of optimal control to provoke robots to achieve robust jumping motions. Robots such as \textit{ANYmal}\cite{animal_01}, \textit{SALTO-1P}\cite{salpo_02} and \textit{Mini Cheetah}\cite{03} can perform remarkable jumping motions and navigating in the local environment. A traditional method simplifies the robot model as a 2-D planar model by combining left and right legs to accelerate reference trajectory optimization speed. Model Predictive Control (MPC)\cite{04}, Whole-Body Impulse Control (WBIC)\cite{05}, and Virtual Model Control (VMC)\cite{06} are implemented to track the reference trajectory. The MIT \textit{Cheetah} 2 is able to autonomously jump over obstacles up to 40 cm in height during bounding gait through Nonlinear MPC\cite{07}. Except for single rigid body models, with kino-dynamic and novel torque-speed limitation constraints, the \textit{Mini Cheetah} can reliably produce successful aerial motions such as flips and barrel rolls\cite{08}. The cost on the order of seconds rather than milliseconds confines the performance of the kino-dynamic planners to generate a motion plan, which implies the real-time limitation of running jumping that starts from the non-static initial position. Though our previous work\cite{09},\cite{10} based on offline Differential Evolution (DE) and online evolutionary algorithms with pre-motion library also endow quadrupedal robots with the ability to execute complicated jumping motions, the trajectory optimization result is not reliable in some extreme cases.

Reinforcement Learning (RL) is another broad way to achieve accurate and aggressive quadrupedal jumping motions. In \cite{11}, with policies considering the robot's total power limits and torque-speed relationships, Unitree A1\cite{11.5} achieves aggressive and accurate jumping motions. Cat-like jumping exploits the possibility of adjusting robot body gestures by jumping motions in a low gravity environment \cite{12}. However, few researchers attach great importance to using a single policy to make robots perform complicated or omnidirectional jumping motions.

For real-time omnidirectional jumping frameworks controlled by a microcontroller unit, the robot \textit{Moobot}\cite{13} is capable of traversing different platforms from all directions. However, the insect-scale robot cannot jump consecutively and must be reloaded by hand after each jump. As for quadruped robots, a hierarchical planning and control framework\cite{14} is proposed to enable \textit{Mini Cheetah} to traverse complex multi-layered terrain. A novel high-level jump selection controller is implemented to ensure the most robustness guarantees. In our work, enabling quadruped robots to perform omnidirectional jumping consecutively, we proposed a framework that minimizes the computing time of the trajectory planner and the reference tracking controller to improve the agility of quadruped robots significantly. Due to analytic solutions, the minimum-jerk trajectory planner efficiently generates a CoM reference path in the jumping phase with numerical stability and dynamic compatibility. Aiming to track reference motions, an intuitive virtual model controller is deployed to compute the force of feet with low computational resources. Then, in the landing phase, a whole-body controller guarantees the CoM inside the support polygon of four feet, ensuring the stability and compatibility of the robot's body.

This paper makes the following contributions:
\begin{itemize}
    \item We propose an omnidirectional jumping framework that generates and tracks aerial motions for quadrupedal robots, made up of a minimum jerk CoM trajectory planner, a VMC reference tracking controller, and a WBC landing controller.
    \item The average cost of generating trajectory and computing tracking motor torque commands to perform omnidirectional jumping motion is within 50 us, implying the remarkable real-time performance of the framework.
    \item  The efficacy of the framework is verified on open-source \textit{Mini Cheetah}\cite{Katz_23}, and the robot succeeds in generating multiple reference CoM trajectories and performing omnidirectional jumping behaviors consecutively (see Fig. 1)
\end{itemize}

The remaining content of this paper is structured as follows. Section \ref{Sec.2} briefly introduces the models of robot and omnidirectional jumping. Section \ref{sec.3} details the formulation of the minimum jerk trajectory planner and the VMC controller, and section \ref{sec.4} is the implementation details and experiments, including the hardware setup, real-time performance, and, verification of the omnidirectional jumping.

\section{Models and dynamics} \label{Sec.2}
\subsection{Robot Model}
The reduced-order dynamic model of jumping motion treats the robot as a single rigid body (SRB) with a specified moment of inertia. The robot state $\bm x$ can be written as:
\begin{subequations}
\begin{eqnarray}
& \bm x:=\lbrack {\bm P_{com}^T} \quad {\bm \Theta^T}\quad {\bm V_C^T} \quad  {\bm \omega^T_B}\rbrack^T \in \mathbb{R}^{12}\label{q_joint_2}\\
&{\bm Q}:=[{\bm q_{i}} \quad {\dot{\bm q}_{i}}] \in \mathbb{R}^{24} \label{q_joint_1}
\end{eqnarray}
\end{subequations}
Where ${\bm P_{com}} \in \mathbb{R}^{3}$ is the position of the robot's body center of mass (CoM) with respect to (w.r.t.) inertial frame (see Fig.\ref{robot_coordinate}); ${\bm \Theta =[\psi \ \phi \ \theta]}$ represents the Euler angles of the robot's body; ${\bm V_C \in \mathbb{R}^{3}}$ is the velocity of the CoM. ${\bm \omega}_B \in \mathbb{R}^{3}$ is the angular velocity of CoM w.r.t. the robot frame $\{B\}$. $\bm q_i \in \mathbb{R}^3$ and $\bm \dot{\bm q}_i \in \mathbb{R}^3$ are the joint angles and velocities of each leg. $i$ is the number of feet. The ground reaction force (GRF) $\bm f_{i} \in \mathbb{R}^3$ at each contact point consists of the dynamic system control input ${\bm u}:=[{\bm f_{i}}]$. ${\bm r_i}$ is the vector from CoM to the robot foot. Then, The linear acceleration of CoM and the angular acceleration of the base are shown:
\begin{subequations}
\begin{align}
m\ddot{\bm P}_{com}&={\sum_{i=1}^{n_c} \boldsymbol{f}_i}- {\bm g} \label{acc}\\
\frac{\mathrm{d}}{\mathrm{d} t}({\bm I_B \bm \omega_B})& = \sum_1^{n_c}\bm{r}_i\times \bm f_i +{\bm \omega_B}\times{\bm I}_B{\bm \omega}_B 
\label{rcc},
\end{align}
\label{centroidal_model}
\end{subequations}
where ${\bm g} \in\mathbb{R}^3$ represents gravitational acceleration. $n_c$ represents the number of contacts. ${^B\bm{I} } \in \mathbb{R}^{3\times3}$ is the robot's rotation inertial tensor, which is assumed as a constant in this work, $\textrm{diag}(^B\bm{I})=[0.07,0.26,0.242]^T$.
\begin{center}
\vspace{-0.5cm}
\begin{figure}[htb]
\centering
\includegraphics[width=3in]{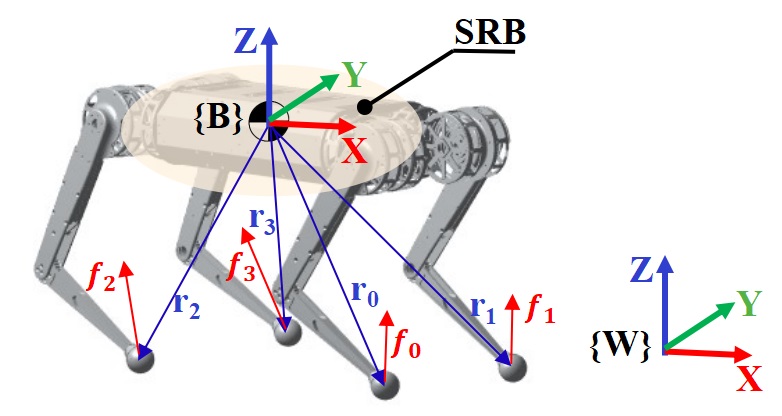}
\caption{A model of a single rigid body (SRB) utilized in the framework for VMC. The blue arrow represents the CoM to the plantar position vector, while the red arrow represents the Ground Reaction Forces (GRFs).}\label{robot_coordinate}
\end{figure}
\end{center}
\vspace{-0.7cm}
\subsection{Omnidirectional Jumping Model}
\begin{figure}[htb]
\centering
\vspace{-0.2cm}
\includegraphics[width=3.2in]{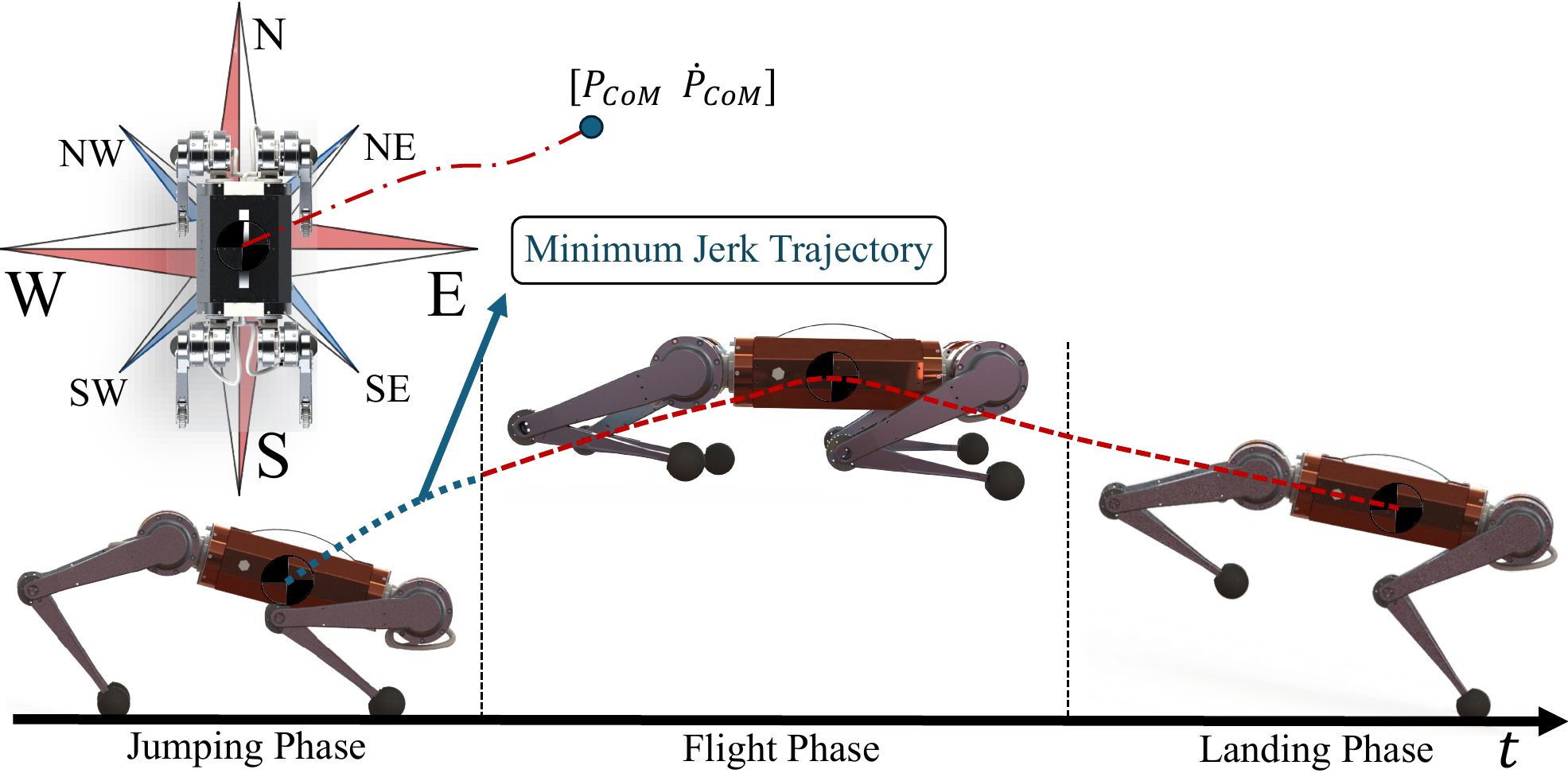}
\caption{The Jumping phase, Flight phase, and Landing phase consist of the omnidirectional jumping motion.}\label{omnidirectional}
\vspace{-0.3cm}
\end{figure}

Omnidirectional jumping significantly increases quadruped robots' capability to access terrains instead of traditional assumptions that robots jump in 2D. The motion of omnidirectional jumping comprises three phases: preparing, flight, and landing (see Fig. \ref{omnidirectional}). In the preparing phase, given desired $[\bm P_{CoM} \ \bm{\dot{P}}_{CoM} \ \bm{\ddot{P}}_{CoM}]$, mini jerk planner and motor torques generate a smooth CoM trajectory are calculated the motor torque commands $\tau_{cmd}$ to track the reference path. In the flight phase, a proportional and derivative controller ensures the preparation of the robot landing. In the landing phase, given reference GRFs and desired base angles, a whole-body controller maintains the stability of the robot's base regardless of the impact of the ground on the robot's legs.

\section{Framework} \label{sec.3}
\begin{figure*}[t]
\centering
\vspace{0.2cm}
\includegraphics[width=6.5in]{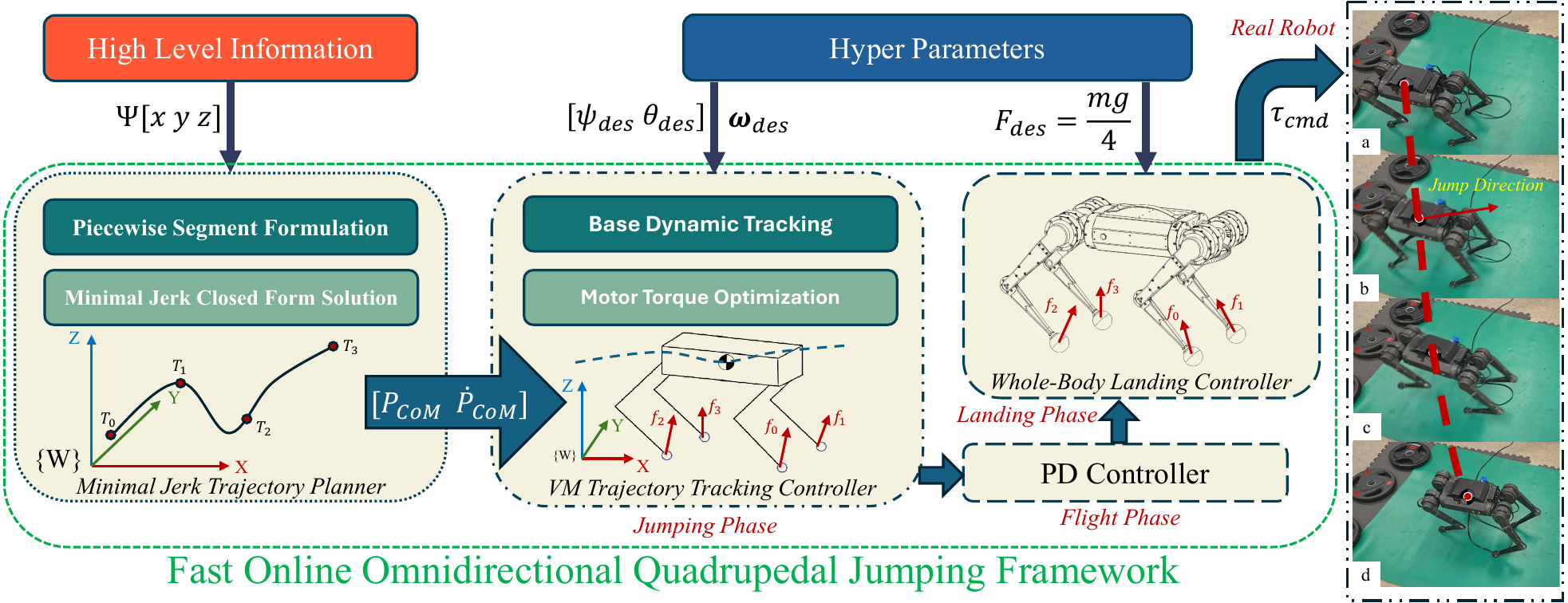}
\vspace{-0.3cm}
\caption{The overview of the fast online omnidirectional quadrupedal jumping framework, made of 3 primary components.}
\label{jump_framework}
\vspace{-0.3cm}
\end{figure*}

This section introduces the online omnidirectional jumping framework with low computational requirements, comprising three primary components: a minimum jerk trajectory generator, a virtual model controller, and a whole-body landing controller. With an analytic solution of relative coefficients, the five-order polynomial trajectory generator rapidly generates a smooth and dynamic feasible path for the CoM of robots based on its current states. Through optimizing output GRFs, the VMC enables the robot to track the CoM reference trajectory using a quadratic program and a Cartesian PD controller. Finally, given reference GRFs and reference base Euler angles, the whole-body landing controller ensures the stability of the robot's landing, and a PD motor torque controller is implemented to improve the robot's performance of tracking references in the landing phase. An overview is shown in Fig \ref{jump_framework}. The first subsection shows details of the mini-jerk trajectory planner, the second section reveals the process of generating motor torque commands by VMC, and the last subsection briefly introduces the WBC and motor torque control methods in the landing phase.

\subsection{Minimum Jerk Trajectory Planner} 
The minimum jerk/snap algorithm has been extensively utilized in UAV systems\cite{Mellinger_15}. UAVs possess unique dynamic features that satisfy the criteria of differential flatness, demonstrating that UAV control can be used for trajectory generation. When the UAV system employs mini-jerk trajectory creation, the UAV experiences minimal thrust, while using mini-snap results in the minimum differential thrust. For the quadrupedal robot, considering the SRB model, when all four feet are in contact with the ground, only the force in the Z direction and the acceleration of the CoM are taken into account, similar to trajectory generation in a UAV system. The system with a fixed inclination satisfies the differential flatness property even if the force direction does not align perfectly with the vector leading to the center of mass along the sole of the quadruped robot\cite{Mu_2019}.

Currently, the trajectory produced by the mini-jerk allows for the generation of relatively reasonable GRFs in a short amount of time. (\ref{acc}) the connection between the CoM acceleration and the contact force helps create a smooth trajectory for the quadruped robot's jumping motion. This trajectory is then sent to the VMC for tracking and optimization to achieve a 12-dimensional contact force that meets kino-dynamics requirements.
In this work, reference Euler angles 
 and angular velocities are selected as hyperparameters, and the trajectory outputs are chosen as:
\begin{align}
s(t) = [x \quad y \quad z] = \bm \psi
\end{align}
To make a quadruped robot well track the space the flatness outputs, the polynomial order of the smooth trajectory is selected as five, and it is convenient to formulate it as three-piecewise segments:
\begin{subequations}
\begin{align}
s(t) &= \left\{
\begin{aligned}
    s_1(t) = \sum & _{i=0}^{5}a_{1,i}t^i \quad T_0 \leq t \leq T_1\\
    s_2(t) = \sum & _{i=0}^{5}a_{2,i}t^i \quad T_1 \leq t \leq T_2 \\
    s_3(t) = \sum & _{i=0}^{5}a_{3,i}t^i \quad T_2 \leq t \leq T_3
\end{aligned} \right. \\
s_j(t)^{'''}  &= \sum_{i\geq3}^5i(i-1)(i-2)t^{i-3}a_{j,i}
\end{align}
\end{subequations}
Where $s(t)$ represents the whole trajectory of CoM, $ s_j(t)$ is each segment's trajectory. $a_{j, i}$ is the coefficient of each 5-order polynomial segment path. $s_j(t)^{'''}$ is the jerk of each segment.

Under the precondition of tracking reference trajectory, smaller reference GRF values in (\ref{acc}) ensure motors' accuracy of torque and speed. Therefore, the optimization program of figuring out a trajectory equipped with the minimal acceleration reference can be formulated as follows:
\begin{subequations}
\begin{eqnarray}
J_j(T) &=& \int_{T_{j-1}}^{T_j}(s_j(t)^{'''})^2dt = \bm p_j^T\bm Q_j \bm p_j \\
s.t.  \quad s_j^{k}(T_j) &=& \bm \psi_{T,j}^{k} \label{constrant_1}\\
s_j^{k}(T_j) &=& s_{j+1}^{k}(T_j) \label{constraint_2} 
\end{eqnarray}
\end{subequations}
where (\ref{constrant_1}) is the derivative constraint for one polynomial segment, (\ref{constraint_2}) is the continuity constraint between two segments, and k is the derivative order. Then the solution of this problem can be solved in closed form ref to \cite{Richter_17}:
\begin{subequations}
\begin{eqnarray} \label{cost_closed_form}
J &=& \left[\begin{array}{c}
     \bm a_f  \\
     \bm a_p 
\end{array}\right]^T\underbrace{\left[\begin{array}{cc}
     \bm R_{ff} & \bm R_{fp}  \\
     \bm R_{pf} & \bm R_{pp}
\end{array}\right]}_{\bm R}\left[\begin{array}{c}
     \bm a_f  \\
     \bm a_p 
\end{array}\right]\\ 
  &=&\bm a_f^T\bm R_{ff}\bm a_f + \bm a_f^T\bm R_{fp}\bm d_p+\bm a_p^T\bm R_{pf}\bm a_f + \bm a_p^T \bm R_{pp}\bm a_p \label{cost_closed_form_2}
\end{eqnarray}
\end{subequations}
where $\bm a = \bm C^T
\left[\bm a_f \quad \bm a_p   \right]^T$, $\bm C$ is the selecting matrix that divides $\bm a$ to free $\bm a_f$ and constrained $\bm a_p$. $\bm R=\bm C \bm M^{-T}\bm Q \bm M^{-1} \bm C^T$, $\bm M$ is the decision variable mapping matrix. The derivative of (\ref{cost_closed_form_2}) gives the solution: $\bm a_p^*=-\bm R_{pp}^{-1}\bm R^T_{fp}\bm a_f$.

Given a final position target w.r.t body frame, piecewise trajectories depend on the piecewise time allocation. We assume the robot body's trapezoidal velocity to simplify the time allocation process. Every piece accelerates to max velocity, keeps its velocity, and decreases to $0m/s^2$.  
\subsection{Virtual Model Trajectory Tracking Controller}
The last subsection generates the robot's base's reference position, velocity, and acceleration. Given the reference acceleration and reference angular acceleration, the GRFs can be solved rapidly with a virtual model controller. The virtual components producing virtual forces do not exist in the real world. Choose spring and damper as the robot's virtual components, and the demonstration is shown as Fig. \ref{jump_framework} and the relationship between $\ddot{\bm P}_{com}$, $\dot{\bm \omega}_B$ and the reference trajectory from minimum jerk is:
\begin{subequations}
    \begin{align}
        \ddot{\bm P}_{com}^r &=\bm k_p^p(\bm P_{com}^r-\bm P_{com}) + \bm k_d^p(\dot{\bm P}_{com}^r-\dot{\bm P}_{com}) \\
        \dot{\bm \omega}^r_B &=\bm k_p^{\omega}\bm e_R + \bm k_d^{\omega}(\bm\omega^r_B-\bm\omega_B)
    \end{align}
\end{subequations}
where $\ddot{\bm P}_{com}^r$ is the reference acceleration, and $\dot{\bm P}_{com}^r$, $\bm P_{com}^r$ are reference base's velocity and position separately from the trajectory planner. $\bm k_p^p \in \mathbb{R}^{3\times 3}$ and $\bm k_d^p \in \mathbb{R}^{3\times 3}$ are the proportional and derivative gains matrices. $\bm \omega_B^r$ is a constant, and $\bm e_R$ is the error function for the rotation matrix, given by \cite{bullo_18} as $e_R=log(\bm R_r^T\cdot \bm R)^v$. $\bm k_p^{\omega}$ and ${\bm k_d^{\omega}}$ are proportional and derivative gain matrices. The reason for not considering the acceleration generated from the trajectory planner is to get a better trajectory tracking performance while robots prepare to jump. The accelerations produced by the virtual model promote the robot's dynamic efficiency and tracking performance with rarely any additional computational cost.

Rewrite (\ref{acc}) and (\ref{rcc}) in matrix formulation\cite{Focchi_19}:
\begin{equation} \label{vmc model}
    \underbrace{\left[\begin{array}{ccc}
        \mathbb{I} & \cdots & \mathbb{I}\\
        \bm r_1\times & \cdots & \bm r_4\times
    \end{array} \right]}_{\bm M \in \mathbb{R}^{6\times 12}}\underbrace{\left[ \begin{array}{c}
         \bm f_1  \\
         \vdots  \\
         \bm f_4 
    \end{array}
    \right]}_{\bm f\in \mathbb{R}^{12\times1}}=\underbrace{\left[\begin{array}{c}
       m\ddot{\bm P}_{com} + \bm g \\
       \bm I_B \ \bm \dot{\omega}_B   
    \end{array}\right]}_{\bm N \in \mathbb{R}^{6 \times 1}}
\end{equation}
With known Cartesian and angular accelerations of the base, the constraint of (\ref{vmc model}) is 6-dimension, but $f$ is 12-dimension. Thus, an optimization program can be formed as follows:
\begin{subequations}
    \begin{align}
        \bm f= \mathop{\arg\min}\limits_f(\bm M \bm f- & \bm N)^T\bm Q(\bm M \bm f-\bm N)+(\bm J_c\bm f)^T\bm R \bm J_c\bm f \label{vmcQP}\\
        s.t. \quad \quad &\bm c_l < \bm{Gf} < \bm c_u \label{friction_constraint}   
    \end{align}
\end{subequations}
Where (\ref{friction_constraint}) is the friction cone constraint of contact feet, whose matrices are shown as:
\begin{eqnarray}
\bm G_i=\left[\begin{array}{c}
     -(\mu_i\bm n_i)^T  \\
     (\mu_i\bm n_i)^T \\
     \bm n_i^T
\end{array}\right],\bm c_l=\left[\begin{array}{c}
     -\infty  \\
     0 \\
     \bm f_{min_i}
\end{array}\right], \bm c_u=\left[\begin{array}{c}
     0  \\
     \infty \\
     \bm f_{max_i}
\end{array}\right]
\end{eqnarray}
$\mu_i \in \mathbb{R}$ is the friction parameter between the contact point and the ground. $\bm n_i \in \mathbb{R}^3$ is the directional normal vector to the ground. $\bm Q \in \mathbb{R}^{6\times 6}$ and $\bm R \in \mathbb{R}^{6\times 6}$ is weight matrices and the term $(\bm J_c\bm f^T)\bm R \bm J_c\bm f$ penalizes large motor torque output when robot's base tracking the references. $\bm J_c\in \mathbb{R}^{6 \times 12}$ is the contact Jacobian Matrix. $\bm c_l\in \mathbb{R}^{20 \times 1}$ is the lower bound of friction cone constrain and $\bm c_u \in \mathbb{R}^{20 \times 1}$ is the upper bound. $\bm G \in \mathbb{R}^{20 \times 12}$ is the mapping matrix.

The \ref{vmcQP} is a quadratic optimized problem and can be solved efficiently by QuadProg++\cite{Goldfarb_20, quadprog_pp}. The motor torque command is calculated by:
\begin{equation}
\begin{aligned}
\bm \tau_{ff} &=-\bm S_f\bm J_c\bm f \\
\bm \tau_{cmd} &= \bm{\tau_{ff}} + \bm k_{cp}(\bm p^r_f - \bm p_f) + \bm k_{cd}({\bm{\dot{p}}^r_f - \bm \dot{\bm p}_f}) 
\end{aligned}
\end{equation}
where $\tau_{ff} \in \mathbb{R}^{12}$ is the motor feed-forward torque vector and $\bm S_f \in \mathbb{R}^{12\times 12}$ is the selecting matrix. $\bm \tau_{cmd}\in \mathbb{R}^{12}$ is the motor command torque. $\bm k_{cp} \in \mathbb{R}^{12 \times 12}$ and $\bm k_{cd} \in \mathbb{R}^{12\times 12}$ are the Cartesian gain matrices separately. $\bm p_f^r$ and $\bm{\dot{p}}_f^r$ are the reference foot position and velocity vectors. $\bm{p}_f$ and $\bm{\dot{p}}_f$ are the actual foot position and velocity vectors.

\subsection{WBC Landing Controller}
As a quadratic program\cite{Fahmi_22}, the whole body controller with a controlling frequency of 500Hz ensures the robot's base stability in the landing phase. When the robot touches the ground, the summation of reference GRFs equals its gravity, and the reference base angles are all zeros. In this way, the robot can land steadily and safely. The whole framework algorithm is shown as algorithm \ref{algo_omni}. The motor torque commands from the whole-body controller come from as follows:
\begin{equation}
    \bm \tau_{cmd} = \bm{\tau_{ff}} + \bm k_p(\bm q^r - \bm q) + \bm k_d({\bm{\dot{q}}^r - \bm \dot{\bm q}})
\end{equation}
where $\bm k_p$ and $\bm k_d$ are the PD gains matrices. $\bm q^r$ and $\bm{\dot{q}}^r$ are the reference motor position and velocity vectors. $\bm q$ and $\bm{\dot{q}}$ are the actual motor position and velocity vectors.

\begin{algorithm} 
\caption{The Framework Algorithm}
\label{algo_omni} 
\SetKwInOut{Input}{input}\SetKwInOut{Output}{output}
	\Input{$ \bm P_{\textrm{com}}^{\textrm{e}},\bm{\dot P}_{\textrm{com}}^{\textrm{e}}, \bm{\ddot P}_{\textrm{com}}^{\textrm{e}}, \bm \omega^{\textrm{e}}, \psi^{\textrm{e}}, \phi^{\textrm{e}}, k$}
	\Output{${\bm \tau_{\textrm{cmd}}}$}
	 \BlankLine 
      \While{$j < k$}
      {
        $s_j(t)=\sum_{i=0}^{5}a_{j,i}t^i \quad T_j \leq t \leq T_{j+1}$\;
        $s_j(t)^{'''} = \sum_{i\geq3}^5i(i-1)(i-2)t^{i-3}a_{j,i}$\;
      }
      $\bm a_p^*=-\bm R_{pp}^{-1}\bm R_{fp}^T\bm a_f$\;
    
      $\ddot{\bm P}_{com}^r =\bm k_p^p(\bm P_{com}^r-\bm P_{com}) + \bm k_d^p(\dot{\bm P}_{com}^r-\dot{\bm P}_{com})$\;
    
      $\dot{\bm \omega}^r_B =\bm k_p^{\omega}e_R + \bm k_d^{\omega}(\bm\omega^r_B-\bm\omega_B)$\;
    
      \While{$n < maximum \ iterations$}{
        {$\min_f \ (\bm M \bm f-\bm N)^T\bm Q(\bm M \bm f-\bm N)+(\bm J_c\bm f)^T\bm R \bm J_c\bm f$}
      
        {\If{$\bm f$ $<$ \textrm{Tolerance}}{\textrm{Terminate}}}
      }
      $\bm \tau_{cmd}=-\bm S_f \bm J_c \bm f $
\end{algorithm}
\DecMargin{1em} 
\section{Implementation Details and Experiments} \label{sec.4}
This part focuses on the hardware implementation specifics of the online omnidirectional jumping framework and experiments conducted using the improved open-source \textit{Mini Cheetah}\cite{Katz_23}. Specifying hardware setup, we first offer solving time data of a particular omnidirectional leaping framework to illustrate the framework's capacity to function in real time on the robot. We showcase the framework's capacity to facilitate Mini-Cheetah executing omnidirectional jumps, such as front-left and rear-right jumps. Finally, comparing the robot performance between the minimum jerk trajectory planner and minimum snap trajectory planner illustrates the disparities between them.
\subsection{Hardware Setup and Real-Time Performance}
The framework developed with C++ is deployed in an enhanced Mini-Cheetah equipped with an Intel NUC i3-8145U@2.1G Hz. The trajectory planning, VMC, and whole-body controllers run on the NUC in real-time. Motor data is collected at 1000 Hz through a New-design USB to Can board. The Lightweight Communications and Marshalling package (LCM) is used for the asynchronous communication between the simulation, hardware bridge, and data collecting. 

Hyper-parameters are shown in Tab. \ref{hyper-parameters}, including hyperparameters of mini jerk trajectory planner and hyperparameters in VMC.

To verify the real-time performance of the fast omnidirectional jumping framework, with the random input reference end CoM position, velocity, and acceleration w.r.t the inertial frame, we record the time-consuming summation of the minimum jerk trajectory planner and VMC controller, which starts from the reference inputting and comes to an end with the generation of motor torque commands. Several time measurements are repeated on actual robots, and the average consuming time is compared with our previous work, the evolutionary-based jumping framework\cite{10} and the offline jumping framework\cite{09}. 
\begin{table}[t]
\vspace{0.3cm}
\caption{HYPERPARAMETERS\label{hyper-parameters}}
\centering
\renewcommand\arraystretch{1.3}
\scalebox{0.9}{
\begin{tabular}{l|l|l}
\hline
Parameters & Symbol &Values \\
\hline
\multicolumn{3}{c}{\textbf{Minimum Jerk}}\\
\hline
End Position Range & $\bm{P}_{CoM}[m]$ & $[\pm 0.15, \pm 0.1, \pm 0.05]$\\
End Velocity Range & $\bm{\dot}{P}_{CoM}[m/s]$ & $[\pm 0.3,\pm 0.16, [1.5,3.5]]$\\
End Acceleration Range & $\bm{\ddot{P}}_{CoM}^z[m/s^2]$ & $[\pm 20,\pm 20, [10,40]]$\\
\hline
\multicolumn{3}{c}{\textbf{VMC}}\\
\hline
Friction Coefficient & $\mu$ & 0.5 \\
Weights for CoM Wrench & $\bm Q$ & $diag(1,1,10,20,10,25)$ \\
Weights for Torque Wrench& $\bm R$ & $diag(5,50,2)10^{-5}$ \\
CoM Proportional Gains & $\bm k_{p}^p$& $diag(1070,1070,1070)$ \\
CoM Derivative Gains & $\bm k_{d}^p$& $diag(12,12,10)$ \\
Attitude Proportional Gains & $\bm k_{p}^{\omega}$& $diag(800,800,800)$ \\
Attitude Derivative Gains & $\bm k_{d}^{\omega}$& $diag(20,10,20)$ \\
GRFs Minimum & $\bm f_{min}[N]$ & 5 \\
GRFs Maximum & $\bm f_{max}[N]$ & 250 \\
Cartesian Proportional Gains & $\bm k_{cp}$ & $diag(0,0,0)$ \\
Cartesian Derivative Gains & $\bm k_{cd}$ & $diag(15,15,15)$ \\
\hline
\end{tabular}
}
\end{table}
As Tab. \ref{tab:algorithm_com} demonstrated, the average time resource required for trajectory planner and VMC is around 50 microseconds on NUC, depicting well the real-time efficacy of the framework. 
\begin{table}[t]
\caption{JUMPING ALGORITHM COMPARISON\label{tab:algorithm_com}}
\centering
\begin{tabular}{c|c}
\hline
Algorithm & Average Time(s) \\
\hline \hline
\textbf{ This work}  & \textbf{0.00005}\\
\hline
Evolution-Based Jumping Framework\cite{10} & 0.14 \\
\hline
Offline Jumping Framework\cite{09} & 65 \\
\hline
\end{tabular}
\end{table}

\subsection{Ominidirectional Jumping}
\begin{figure}[h]
\vspace{0.3cm}
\centering
 \subfigure[]{
    \label{four_feet_pre} 
    \includegraphics[width=3.2in]{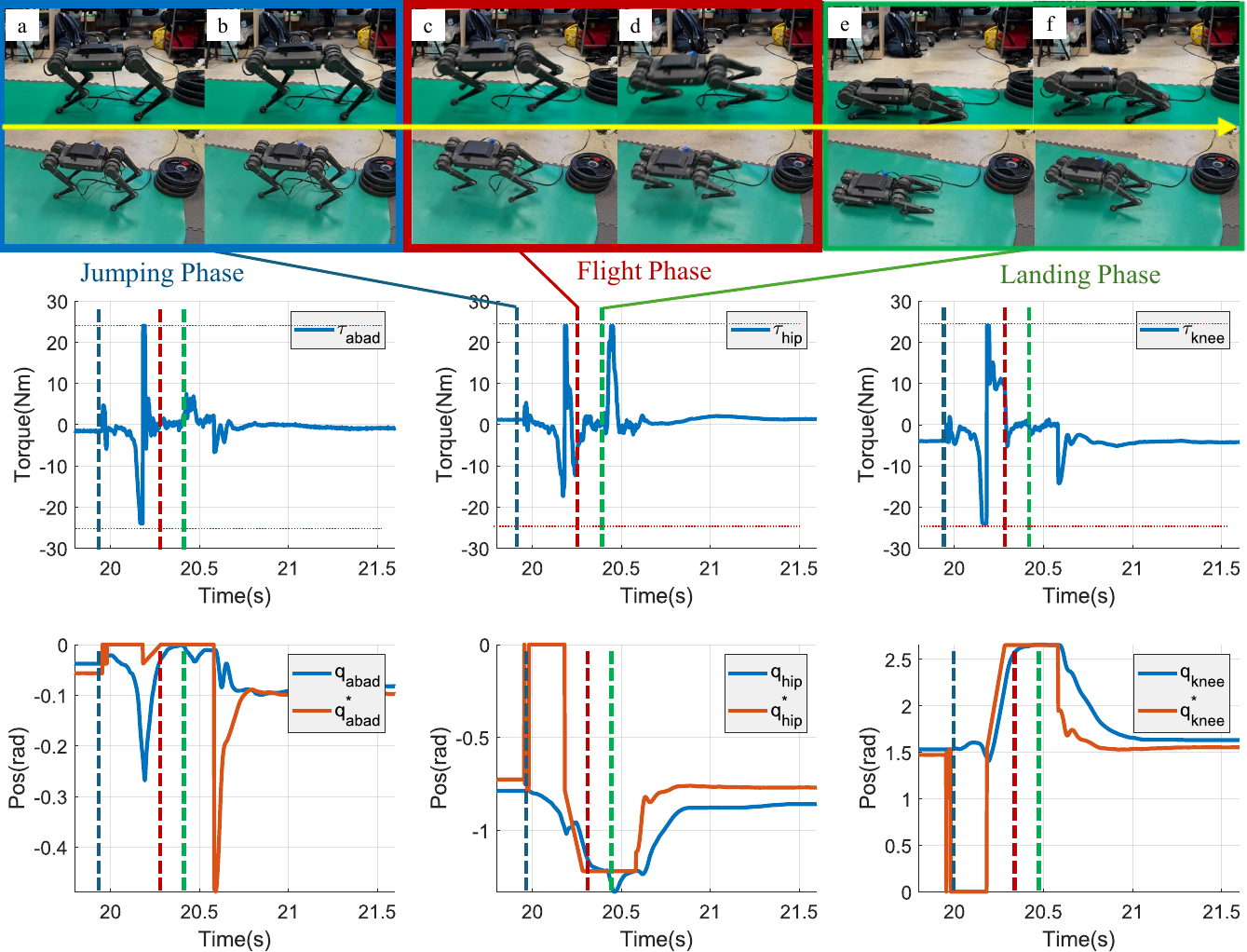}}
 \subfigure[]{
    \label{four_feet_pre} 
    \includegraphics[width=3.2in]{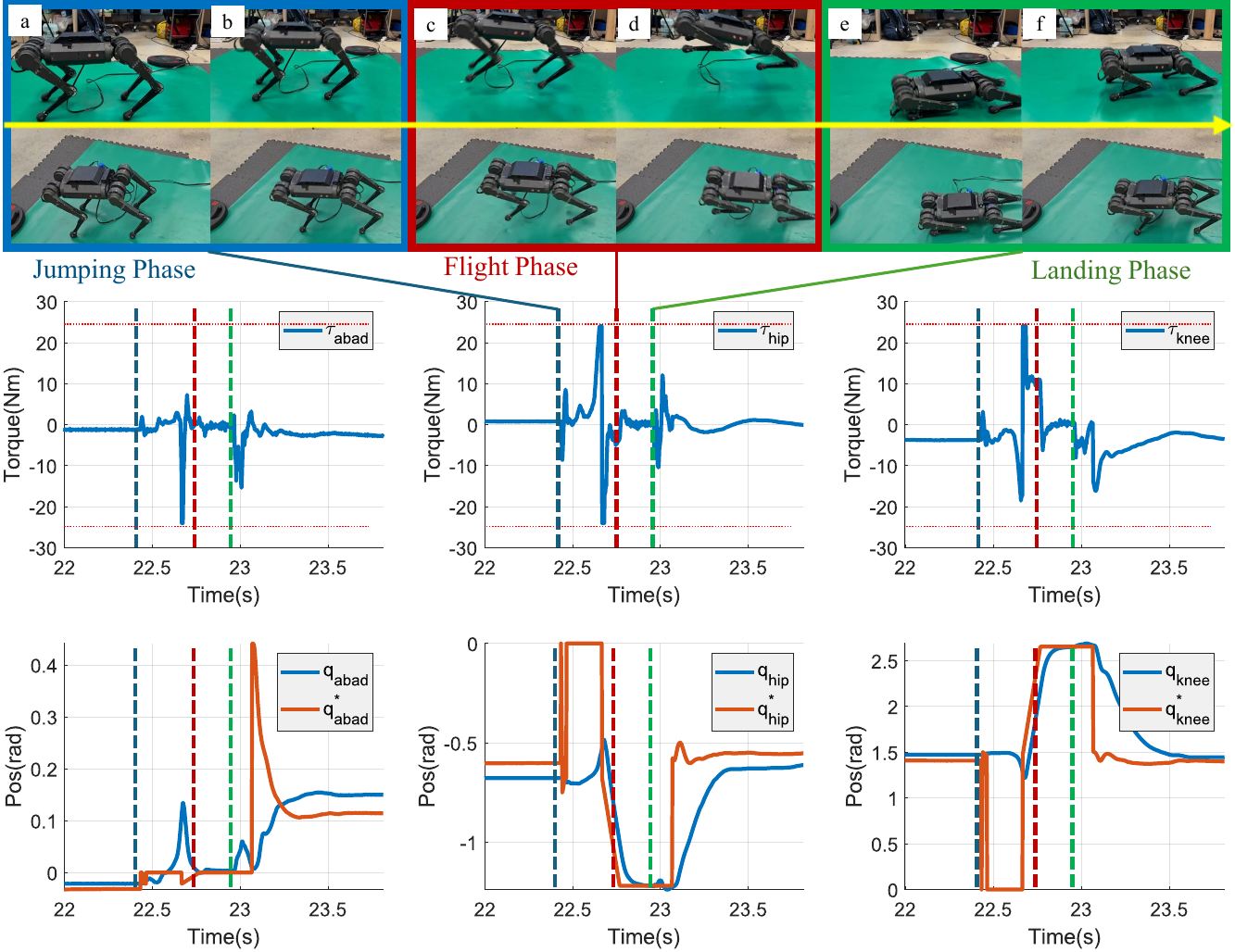}}
\vspace{-0.3cm}
\caption{Mini-Cheetah deployed with the omnidirectional jumping framework performs two jumping motions. Data charts in Fig. (a) and Fig. (b) are data records of the rear-right leg in front-left jumping and rear-right jumping, respectively. The blue line represents the start of the jumping phase, the dark red line indicates the start of the flight phase, the green line is the beginning of the landing phase, and the red dot lines represent the upper and lower bound of the torque limit with an absolute value of 24 Nm. }
\label{perform_jump}
\vspace{-0.3cm}
\end{figure}

Made consisting of the jumping phase, flight phase, and landing phase, the omnidirectional jumping framework is capable of enabling Mini-Cheetah to jump in any direction w.r.t its base frame, such as the front side, the left side, the rear side, and angles between them. Fig. \ref{perform_jump} illustrates the robot's ability to perform omnidirectional jumping well. 
As the figures show, the period between the start of the jumping phase and the beginning of the flight phase is within 0.5 seconds. The maximum of motor torques in the jumping phase is around 22 Nm. Because of the imprecision of the SRB model in dynamics, motor torques generated by the VMC are smaller than the actual desired torques. Thus, a Cartesian PD controller is implemented to improve the performance of foot position and speed tracking in the jumping phase, conducting a better jumping motion. The front-left jumping direction and the rear-right jumping direction are selected intentionally to show the robot's potential capability of performing omnidirectional jumping, which matches the result of the experiments.

\section{CONCLUSIONS}
In conclusion, this paper introduces a new omnidirectional jumping framework for quadruped robots, leveraging Virtual Model Control (VMC) and Minimum Jerk principles. This framework is distinguished by its rapid execution, achieving microsecond levels with impressive speed. By employing a 5th-order polynomial for trajectory generation alongside defined constraints for continuity, the minimum-jerk trajectory method proves highly effective for this application. Implementing a VMC-based trajectory tracking controller, utilizing PD control for centroid trajectory tracking and GRFs optimization, has shown remarkable efficiency, and in real-world jumping experiments, the solution time for trajectory tracking reached approximately 30 us, with an additional 20 us for trajectory generation. This significantly enhances the optimization time to 50 us, marking a substantial improvement over previous methods that relied on the differential evolution algorithm. Such computing speed is critical for fulfilling the real-time requirements of emergency obstacle avoidance tasks in robotic applications.

Looking ahead, our future work will explore improved time allocation techniques and enhanced precision in jump control, particularly in conjunction with external localization systems. We also aim to address the challenges observed with the current impulse-based WBC landing controller\cite{05}, which sometimes obtained failure results in landing phases. A primary focus will be developing a new landing controller that effectively manages horizontal and vertical speeds to preserve landing posture. 
 

\addtolength{\textheight}{0cm}   









\end{document}